\documentclass[conference]{IEEEtran}
\IEEEoverridecommandlockouts
\usepackage{cite}
\usepackage{amsmath,amssymb,amsfonts}
\usepackage{algorithmic}
\usepackage{graphicx}
\usepackage{textcomp}
\usepackage{xcolor}
\usepackage{subcaption}
\usepackage{url}
\usepackage{hyperref}
\usepackage{booktabs} 
\usepackage{array}

\def\BibTeX{{\rm B\kern-.05em{\sc i\kern-.025em b}\kern-.08em
    T\kern-.1667em\lower.7ex\hbox{E}\kern-.125emX}}
\begin{document}

\title{Learning to Optimally Dispatch Power: Performance on a Nation-Wide Real-World Dataset  
}

\author{\IEEEauthorblockN{
Ignacio Boero\IEEEauthorrefmark{1}\IEEEauthorrefmark{2},
Santiago Diaz\IEEEauthorrefmark{1},
Tom\'as V\'azquez\IEEEauthorrefmark{1}, 
Enzo Coppes\IEEEauthorrefmark{1}\IEEEauthorrefmark{3},
Pablo Belzarena\IEEEauthorrefmark{1}
and
Federico Larroca\IEEEauthorrefmark{1}}
\IEEEauthorblockA{\IEEEauthorrefmark{1}Facultad de Ingenier\'ia, Universidad de la Rep\'ublica, Uruguay}
\IEEEauthorblockA{\IEEEauthorrefmark{2}Department for Electrical and Systems Engineering,
University of Pennsylvania, USA}
\IEEEauthorblockA{\IEEEauthorrefmark{3}Administraci\'on Nacional de Usinas y Transmisiones El\'ectricas (UTE), Uruguay\\
Emails: \{iboero,santiago.diaz,tomas.vazquez,belza,flarroca\}@fing.edu.uy and ecoppes@ute.com.uy}
}

\maketitle

\begin{abstract}
The Optimal Reactive Power Dispatch (ORPD) problem plays a crucial role in power system operations, ensuring voltage stability and minimizing power losses. Recent advances in machine learning, particularly within the ``learning to optimize'' framework, have enabled fast and efficient approximations of ORPD solutions, typically by training models on precomputed optimization results. While these approaches have demonstrated promising performance on synthetic datasets, their effectiveness under real-world grid conditions remains largely unexplored.
This paper makes two key contributions. First, we introduce a publicly available power system dataset that includes both the structural characteristics of Uruguay’s electrical grid and nearly two years of real-world operational data, encompassing actual demand and generation profiles. Given Uruguay’s high penetration of renewable energy, the ORPD problem has become the primary optimization challenge in its power network. Second, we assess the impact of real-world data on learning-based ORPD solutions, revealing a significant increase in prediction errors when transitioning from synthetic to actual demand and generation inputs. Our results highlight the limitations of existing models in learning under the complex statistical properties of real grid conditions and emphasize the need for more expressive architectures. By providing this dataset, we aim to facilitate further research into robust learning-based optimization techniques for power system management.
\end{abstract}

\begin{IEEEkeywords}
Learning to Optimize; Optimal Reactive Power Dispatch; Real-World Power System Dataset
\end{IEEEkeywords}

\section{Introduction}

% {\color{red} Un poco de viru-viru sobre la importancia del problema de despacho de potencia (y reactiva en particular, pero no enfocarse sólo en eso). 

% Decir que han habido varias propuestas para ``learn to optimize'' en muchos ámbitos: telecom, signal processing y por supuesto potencia y el despacho óptimo. Sin embargo, si bien muestran resultados muy alentadores, usan típicamente datos (en parte) simulados. 

% Y bueno, decir que nosotros vamos a hacer la contribución de publicar libremente datos de una red real (tanto la estructura como las demandas). Nuestros experimentos muestran como la alta variabilidad de datos reales resulta en un aprendizaje mucho más desafiante, con resultados claramente inferiores respecto al caso de datos simulados.

% despacho de activa no relevante en Uru.

% }

The Optimal Power Flow (OPF) problem is fundamental in power system operations, aiming to determine the most efficient operating conditions for a power network while satisfying system constraints~\cite{frank2016introduction}. Within this framework, the Optimal Reactive Power Dispatch (ORPD) problem focuses specifically on the management of reactive power to maintain voltage stability and minimize power losses. Effective ORPD is crucial for ensuring system reliability, especially with the increasing integration of renewable energy sources, which introduce variability and uncertainty into the grid~\cite{biswas2019optimal}.

In recent years, the ``learning to optimize'' framework~\cite{chen2022l2o} has emerged as a powerful approach to solving complex optimization problems across various domains, including telecommunications or signal processing~\cite{bala2019deep,randall2023learning,borgerding2017amp}. Similarly, in power systems, machine learning-assisted methods have been developed to address the computational challenges of the OPF and ORPD problems~\cite{owerko2020optimal,pan2023deepopf,owerko2024unsupervised}.
The core idea is to train a learning model on historical optimization solutions --obtained from traditional and costly methods-- so that it can predict (near) optimal solutions efficiently during inference at a significantly lower computational cost.
% The "learning to optimize" framework leverages machine learning models to approximate or accelerate the solutions of complex optimization problems by learning patterns from data and past instances. 
This approach not only enhances computational efficiency but also offers the potential to handle the increasing complexity and uncertainty associated with modern power grids. 
% As the energy landscape continues to evolve, integrating machine learning into optimization frameworks represents a promising direction for advancing the reliability and efficiency of power system management.

Despite these advances, existing algorithms and architectures have been trained and evaluated primarily on synthetic datasets, typically based on IEEE X-Bus systems (see \url{https://icseg.iti.illinois.edu/power-cases/}). While these test cases approximate real-world grid topologies, they lack actual demand data. Consequently, researchers have had to rely on synthetic values for these critical inputs, often sampling them uniformly around predefined reference values.
% machine Learning algorithms are only as good as the data they are trained on. 

% The application of the learning to optimize framework to OPF and ORPD problems involves training models to predict optimal power flow solutions, thereby expediting decision-making processes in power system operations. 

The first major contribution of this paper is to present a freely available power system dataset that includes not only the electrical grid elements (buses, loads, etc.), their key parameters and interconnections, but also nearly two years of real-world operational data, including demands and generated powers. This data corresponds to Uruguay's network, where the significant presence of renewable energy has led to a simplified active power dispatch. Essentially, all available renewable is utilized (prioritizing solar and eolic, supplemented by biomass and hydro), covering approximately 97\% of annual demand~\cite{uruguayxxi}. What little remains is covered with either thermal generation or imported electricity. 
As a result, the ORPD problem has become the most critical optimization challenge in Uruguay’s grid, making it the focus of this study.

Our second main contribution is to measure the impact of using real-world inputs in learning-based solutions to the ORPD problem. In Sec.\ \ref{sec:data} we show how actual demands and generated powers exhibit strongly non-uniform distributions, standing in stark contrast with the synthetic datasets. %traditionally used for evaluating learning methods. 
As we present in Sec.\ \ref{sec:experiments}, these characteristics result in a tenfold increase in the Mean Absolute Error (MAE) in the predicted control variables as well as significant constraints violations when applying the learning models typically used in the literature~\cite{owerko2020optimal}. That is to say, fully training and evaluating the same architecture for either synthetic or historical datasets results in a seriously degraded performance in the latter. This highlights the increased complexity of learning-based ORPD when faced with actual grid conditions and underscores the need for more expressive architectures capable of effectively handling realistic dispatch scenarios. By making our dataset publicly available, we aim to provide a critical resource for advancing research in this direction.
% This illustrates the difficulty of the problem when using actual data, and calls for more expressive architectures to tackle the learning to dispatch problem. We believe that publishing our dataset is the first step towards this goal.
\section{Optimal Reactive Power Dispatch}

% An ever-growing integration of \emph{renewable energy sources}---characterized by rapid and difficult-to-forecast variations---makes the \textit{Optimal Reactive Power Dispatch (ORPD)} problem vital for ensuring both robust and efficient system operation. ORPD entails the management of reactive power using compensators, voltage-controlled generators, or adjustable taps in transformers, thereby delivering solutions that strengthen reliability and enhance overall performance.

\smallskip
\noindent
\textbf{Notation and Model Setup.} We define \(\mathcal{B} = [0,\ldots,N]\) as the set of buses in the network. Each bus can have at most one generator, load or reactive compensator. Hence, we denote:
\[
\mathcal{G} = \{g_i \mid i \in \mathcal{B}\}, \quad
\mathcal{L} = \{l_i \mid i \in \mathcal{B}\}, \quad
\mathcal{R} = \{r_i \mid i \in \mathcal{B}\},
\]
representing the sets of buses including each element type. Within the set of generators, we distinguish between those that allow reactive control through a voltage setpoint (set $\mathcal{G}^{volt}$) and those that provide static reactive power ($\mathcal{G}^{stat}$), where $\mathcal{G} = \mathcal{G}^{volt} \cup \mathcal{G}^{stat}$ and $\mathcal{G}^{volt} \cap \mathcal{G}^{stat} = \emptyset$. 
% \textbf{Notation and Model Setup.} We define \(\mathcal{B} = [0,\ldots,N]\) as the set of buses in the network. Each bus can have at most one unit of each element type. Hence, we denote:
% \[
% \mathcal{G} = \{g_i \mid i \in \mathcal{B}\}, \quad
% \mathcal{L} = \{l_i \mid i \in \mathcal{B}\}, \quad
% \mathcal{R} = \{r_i \mid i \in \mathcal{B}\},
% \]
% representing the sets of generators, loads, and reactive compensators, respectively. Within the set of generators, we distinguish between those that allow reactive control through a voltage setpoint and those that provide static reactive power:
% \[
% \mathcal{G} = \mathcal{G}^{volt} \cup \mathcal{G}^{stat}, 
% \quad
% \mathcal{G}^{volt} \cap \mathcal{G}^{stat} = \emptyset.
% \]
Each of these elements is associated with the following variables:
\[
\begin{aligned}
g^{volt}_i \in \mathcal{G}^{volt} &: \quad v_i^{set}, \quad s^{g,volt}_i = p^{g,volt}_i + j\,q^{g,volt}_i,\\
g^{stat}_i \in \mathcal{G}^{stat} &: \quad s^{g,stat}_i = p^{g,stat}_i + j\,q^{g,stat}_i,\\
l_i \in \mathcal{L} &: \quad s^l_i = p^l_i + j\,q^l_i,\\
r_i \in \mathcal{R} &: \quad s^r_i = j\,q^r_i.
\end{aligned}
\]

Here, \(s\), \(p\), and \(q\) represent apparent, active, and reactive powers, respectively, whereas \(v^{set}\) denotes the voltage setpoint for generators with reactive control. Each bus \(i\) is further characterized by the complex voltage \(v_i\) and the injected power \(s_i\). The voltage \(v_i\) is shared by all elements connected to bus \(i\). The net injected power \(s_i\) at bus \(i\) is given by 
\begin{equation}
s_i = s_i^{g,volt} \;+\; s_i^{g,stat} \;-\; s_i^l \;+\; s_i^r.    
\end{equation}

Let \(\mathcal{E} = \{(i,j) \mid (i,j) \in \mathcal{B} \times \mathcal{B}\}\) denote the set of transmission lines connecting the buses. We adopt the \(\pi\)-model for each line \((i,j)\), which is characterized by
the series admittance \(y_{ij}\), the shunt admittances at bus \(i\) and bus \(j\) (\(y_{\text{sh},ij}\) and \(y_{\text{sh},ji}\) respectively) and an ideal transformer parameter \(t_{ij}\), representing the transformation ratio (such that \(v_i = t_{ij}\,v_j\)). If there is no transformer, \(t_{ij} = 1\).

% \begin{itemize}
%     \item The \emph{series admittance}, \(y_{ij}\).
%     \item The \emph{shunt admittances} at bus \(i\) and bus \(j\), denoted \(y_{\text{sh},ij}\) and \(y_{\text{sh},ji}\), respectively.
%     \item An \emph{ideal transformer} parameter \(t_{ij}\), representing the transformation ratio (such that \(v_i = t_{ij}\,v_j\)). If there is no transformer, \(t_{ij} = 1\).
% \end{itemize}

The power flow along a line \((i,j)\) is expressed through the complex quantities \(s_{i \to j}\) and \(s_{j \to i}\), defined by:
\begin{align}
s_{i \to j} \;=\; \frac{v_i\,v_i^*\,}{t_{ij}^2}y_{\text{sh},ij}^* \;+\; \frac{v_i\,(v_i^* - v_j^*)\,}{t_{ij}^2}y_{ij}^*,
\label{eq2}\\
s_{j \to i} \;=\; \frac{v_j\,v_j^*\,}{t_{ij}^2}y_{\text{sh},ji}^* \;+\; \frac{v_j\,(v_j^* - v_i^*)\,}{t_{ij}^2}y_{ij}^*,
\end{align}

% \noindent
% \textbf{Power Flow Equation.} 
According to Kirchhoff’s Current Law, the total power injected at bus \(i\) must be equal to the sum of power flows from bus \(i\) to its adjacent buses:
\begin{gather}
s_i = \sum_{j \,\mid\, (i,j) \in \mathcal{E}} s_{i\to j} ,\quad\forall i \in \mathcal{B}.
\end{gather}
Using \autoref{eq2} we can redefine the constraint as:
\begin{align}
    &s_i =  \sum_{j \,\mid\, (i,j) \in \mathcal{E}} \Big( \frac{v_i\,v_i^*\,}{t_{ij}^2}y_{\text{sh},ij}^* \;+\; \frac{v_i\,(v_i^* - v_j^*)\,}{t_{ij}^2}y_{ij}^* \Big) &&\forall i \in \mathcal{B}.
\label{eq:ac-flow}
\end{align}
\autoref{eq:ac-flow} is commonly known as the \emph{power flow equation}, and it implicitly determines the bus voltages once all power injections in the network are specified.

% \noindent
% \textbf{Operational Constraints.}
Furthermore, and to ensure reliable system operation, we impose constraints on the magnitudes of the involved variables. For bus voltages, we have that
\begin{align}
v_{\min,i} \;\leq\; &\lvert v_i \rvert \;\leq\; v_{\max,i} ,
&& \forall i \in \mathcal{B}, \label{eq:constraint_ejemplo}\\
\Delta \delta_{ij,\min} \;\leq\; &v_i\,v_j^* \;\leq\; \Delta \delta_{ij,\max},
&& \forall i \in \mathcal{B}, \\
\lvert  v_i \rvert \;=&\; v_i^{set},
&& \forall i \mid g_i^{volt} \in \mathcal{G}^{volt}.
\end{align}
whereas the power flow must not exceed the line's capacity:
\begin{align}
&\lvert s_{i \to j} \rvert \;\leq\; s_{i \to j,\max}, && \forall (i,j) \in \mathcal{E}\cup\mathcal{E}^T.
\end{align}
Finally, any element responsible for reactive compensation must keep its reactive generation within allowable limits:
\begin{align}
q^{g,volt}_{i,\min} \;\leq\; &q^{g,volt}_i \;\leq\; q^{g,volt}_{i,\max}, && \forall i \mid g^{volt}_i \in \mathcal{G}^{volt}, \\
q^{r}_{i,\min} \;\leq\; &q^{r}_i \;\leq\; q^{r}_{i,\max}, && \forall i \mid r_i \in \mathcal{R}.
\end{align}

\noindent
\textbf{Problem definition.} Various objectives can be pursued depending on the priorities of system operators. Common goals include minimizing active power losses in the transmission lines $P_{\text{loss}}$ or maintaining voltage magnitudes within safe levels. In this work, we focus on the former objective, although the methodology can be easily extended to other cases. Thus, the ORPD problem is formulated as:
\begin{equation}\tag{ORPD}
\label{eq:ORPD}
\begin{aligned}
\min_{v_i^{set},\,q_i^r}\,
&\sum_{(i,j)\in \mathcal{E}} \operatorname{Re}\bigl(s_{i\to j} + s_{j\to i}\bigr)\\
\text{s.t.}\, \quad
&\text{Eqs.~(1) -- (11).}
\end{aligned}
\end{equation}

% \noindent
% \textbf{Solution Methods and Challenges.} 
Solving the  ORPD can be extremely complex due to its non-convex nature. Although IPOPT can yield feasible, high-quality solutions, it may be too time-consuming for real-time applications or large-scale scenario analyses. Consequently, machine learning techniques present an attractive option, potentially offering similarly accurate results in a fraction of the time.

\section{Learning to Dispatch}

% {\color{red} La sección 3.5 de la tesis. Realmente no tengo claro si vale la pena hablar de los dos métodos (en particular no sé si vale la pena hablar del unsupervised), más que nada por la falta de lugar.}

In the previous section we introduced the problem to be solved. The variables involved can be categorized into three types: input, control, and implicit variables. The first are fixed for a given instance of the problem, such as demands or generated power. For each node $i$ in the network, we define the input variable as a column vector:
\begin{equation}
\mathbf{x}_{i} = [p^{l}_{i}, q^{l}_{i},p^{g,stat}_{i}, q^{g,stat}_{i}, p^{g,volt}_{i}]^\top,
\end{equation}
%where $p^{\text{dem}}_{i}$ and $q^{\text{dem}}_{i}$ denote the active and reactive power demands, $p^{\text{gen},\text{est}}_{i}$ and $q^{\text{gen},\text{est}}_{i}$ correspond to the active and reactive power generated by static generators, and $p^{\text{gen},v}_{i}$ represents the active power from voltage-controlling generators. 
A value of zero is assigned when a variable is not defined (e.g.\ $p^{l}_{i}=0$ at buses without an attached load).

% Note that for the rest of the variables, there are some that are not directly controllable by the network operator; e.g.\ voltages at loads. We will denote these variables as implicit. 
On the other hand, setpoints of the voltage-controlling generators $v_{i}^{set}$ and the reactive power generated by compensators $q_i^{\text{r}}$ can be controlled directly by the network operator, and as such we will define the control variable at each node as:
\begin{equation}
\mathbf{y}_{i} = [v_{i}^{set}, q_i^{r}]^\top.
\end{equation}
As before, undefined values are set to zero. Finally, the values of the rest of the variables, such as voltages at load nodes and power through the lines, are set implicitly by the input and control variables (for instance through \autoref{eq:ac-flow}), and we will thus denote them as implicit variables.

The ORPD problem thus consists of determining the optimal control variable $\mathbf{Y}^*=[\mathbf{y}_1^*| \mathbf{y}_2^*| \ldots | \mathbf{y}_N^*]^\top$ for a given input $\mathbf{X}=[\mathbf{x}_1| \mathbf{x}_2| \ldots | \mathbf{x}_N]^\top$ such that the resulting implicit variables minimize losses and satisfy the safety restrictions.

A natural approach to obtaining the mapping $\boldsymbol{\Phi}:\mathbb{R}^{N\times 5}\rightarrow \mathbb{R}^{N\times 2}$ such that $\boldsymbol{\Phi}(\mathbf{X})=\mathbf{Y}^*$ is solving the ORPD problem via IPOPT. However, such procedure may prove overly time-consuming under real-time operation, or when repeated calculations are needed (e.g.\ for ``what-if'' scenario analysis). Furthermore, availability of an actual functional expression for $\boldsymbol{\Phi}(\mathbf{X})$ enables applications like sensitivity analysis.

% Here we explore the so-called ``learning to optimize'' framework, where we have a set of pairs $\{(\mathbf{X}_k,\mathbf{Y}^*_k)\}_{k=1,\ldots,K}$ (where $\mathbf{Y}^*$ is computed offline, for instance through IPOPT) and a family of functions parametrized by $\boldsymbol{\theta}$. The objective now becomes finding the best parameter $\boldsymbol{\theta}^*$ so that 
We explore here the ``learning to optimize'' framework, where we approximate $\boldsymbol{\Phi}$ using a parametric model trained on a dataset of optimal solutions $\{(\mathbf{X}_k,\mathbf{Y}^*_k)\}_{k=1,\ldots,K}$ computed offline (e.g.\ using IPOPT). The objective is to find the best set of parameters $\boldsymbol{\theta}^*$ minimizing:
\begin{gather}
    \boldsymbol{\theta}^*=\underset{\boldsymbol{\theta}}{\mathrm{argmin}} \frac{1}{K}\sum_{k=1}^K L\left(\boldsymbol{\Phi}_{\boldsymbol{\theta}}(\mathbf{X}_k),\mathbf{Y}^*_k\right),
\end{gather}
where $L(\cdot,\cdot)$ is a loss function, such as the mean square error (MSE) at those entries that are well-defined (e.g.\ we ignore the second coordinate of $\mathbf{y}_i$ at nodes corresponding to a bus with no reactive power compensator).
% : $\|\boldsymbol{\Phi}_{\boldsymbol{\theta}}(\mathbf{X}_k)-\mathbf{Y}^*\|^2_F$. {\color{red} acá no usamos ninguna máscara? porque las entradas que son siempre cero no deberíamos darle bola en la salida, no?}

The function family $\boldsymbol{\Phi}_{\boldsymbol{\theta}}$ may be a Fully-Connected Neural Network (FCNN) or a Graph Neural Network (GNN). The advantage of GNNs, which in a nutshell is a concatenation of layers each consisting of a graph convolution followed by a non-linearity~\cite{gama2020graphs}, is their ability to naturally incorporate the network topology. For instance, simply adding a node to the network typically requires re-training an FCNN, whereas a GNN can generalize to the new structure~\cite{gama2020stability}.

This supervised approach has demonstrated strong performance on synthetic data~\cite{owerko2020optimal}. Here, we evaluate it on real-world demand and generation data, showing that real data introduces greater complexity compared to synthetic scenarios, necessitating more expressive learning models. Additionally, we revisited the unsupervised learning approach from~\cite{owerko2024unsupervised}, reaching similar conclusions. However, for brevity, those results are omitted.

\section{Contributed Data}\label{sec:data}

% {\color{red} Propongo hablar únicamente de la red uruguaya (y los datos asociados) y no de la IEEE (que por supuesto se debe mencionar). De esta forma se tiene más destaque de una contribución, sino que además ahorramos espacio. Dar algún numerito de la red en sí, pero sobre todo destacar el comportamiento poco homogéneo de los datos (la sección 5.1 de la tesis), que son lo que realmente hacen a que los algoritmos de aprendizaje no anden tan bien. 

% Incluir URL con los datos.}

% A key aspect of evaluating learning-based approaches to the ORPD problem is the availability of representative datasets for training and validation. Most studies rely on synthetic datasets, such as IEEE X-Bus systems, which serve as standardized benchmarks for testing optimization algorithms. However, to explore the applicability of these methods in real-world scenarios, 
% which a detailed model of Uruguay’s electrical network,

We now present the dataset we constructed using several data sources provided by Uruguay's \textit{Despacho Nacional de Cargas} (DNC). The dataset includes both the structural details of the power grid and real operational data for each node, enabling the analysis of learning-based ORPD under realistic network conditions.  %, allowing us to analyze learning-based ORPD under actual network conditions.
 Constructing this dataset was a major undertaking, requiring meticulous efforts to ensure its accuracy and representativeness. To support further research in machine learning for power systems, we have made the dataset—along with the code for the analyses presented in this work—freely available at \url{https://github.com/tomyvazquez/DORAA-UY}.
% \footnote{\url{https://drive.google.com/drive/folders/121s67\_IgW-r39hG-xwHIUI32-\_QIE97X?usp=sharing}}
%
We also considered a synthetic dataset for completeness, enabling direct comparison with prior work. 
%, while the latter provides insight into the behavior of data-driven models when exposed to real operating conditions. 
% This section details the data sources, the process of building the Uruguayan network model, and the key differences between the datasets used.

\noindent\textbf{Real-World Dataset}.
The modeled Uruguayan power system consists of $|\mathcal{B}|=107$ buses, of which $95$ operate at $150$ kV and $12$ at $500$ kV. The network includes $|\mathcal{E}|=156$ transmission lines, where $14$ connect $500$ kV buses, $130$ connect $150$ kV buses and $12$ are transformers that connects each $500$ kV bus with a respective bus of the same name at $150$ kV.
The system features $|\mathcal{G}| = 43$ generators, from which $|\mathcal{G}^{volt}|=15$ are voltage controllers and $|\mathcal{G}^{stat}|=27$ are static generators. Finally, there are $|\mathcal{L}|=55$ loads and $|\mathcal{R}|=6$ reactive power compensators.
All electrical parameters of these elements are also available. 

Furthermore, the dataset includes the generated power for each generator, covering the period from January 1, 2021, to December 31, 2023, with hourly resolution, resulting in a total of 26,280 entries.
It also includes the total power consumed per bus with loads, spanning from March 21, 2021, to October 17, 2023, with a 10-minute resolution, yielding 135,504 data points.
These datasets provide a detailed temporal profile of Uruguay’s power system. %forming the foundation for evaluating learning-based approaches under real network conditions.

\noindent\textbf{Synthetic Data Generation}.
We follow the approach described in~\cite{owerko2020optimal,owerko2024unsupervised} to generate the synthetic dataset. First, we determine the nominal generation and demand values for each bus by computing the time-averaged historical values. Using these nominal values as a baseline, we then sample 10,000 instances, where each feature at every node is drawn from a uniform distribution within $\pm$30
\% of its nominal value. This method ensures that the synthetic dataset captures realistic fluctuations.% while maintaining statistical consistency with observed data.
% To construct the synthetic dataset we proceeded as in~\cite{owerko2020optimal,owerko2024unsupervised}. The first step is to determine the nominal generation and demand values for each bus by averaging their respective historical values over time. Once these nominal values are obtained, the dataset is generated by sampling 10,000 instances where each feature of every node is drawn from a uniform distribution ranging between $\pm$30\%  of its corresponding nominal value. This approach ensures that the synthetic data captures realistic variations.% while maintaining consistency with the patterns observed in the real network.

\noindent\textbf{Generating Output Data}.
For both the real and synthetic datasets, the optimal reactive power dispatch is solved for each input $\mathbf{X}_t$ using the PandaPower library~\cite{pandapower}, where $t$ denotes the timestamp of the extracted data, advancing in hourly increments. 
Since one dataset contains values at an hourly resolution and the other every 10 minutes, we ensured completeness and consistency by selecting data at hourly intervals. This approach allows us to match timestamps across datasets, ensuring alignment. Optimizing through PandaPower determines the bus voltage magnitudes and the reactive power injection from compensators, producing the outputs $\mathbf{Y}^*_t$.

% {\color{red} Acá hay que aclarar cómo se hizo con las dos escalas temporales que hay: una de una hora y otra de diez minutos. Y mencionar quién sería entonces $t$, que hasta ahora no lo puse (usé $k$). }

% \begin{equation}
%     y_{t,i} = [v_{m,i}, q_i^{\text{comp}}]
% \end{equation}.

% By concatenating the results for all buses, the final output vector 
% \begin{equation}
% \mathbf{y}_t = [y_1| y_2| \cdots | y_N]
% \end{equation}
%  is obtained. This completes the dataset construction process.

\begin{figure}
    \centering
    % \begin{subfigure}{0.24\textwidth}
        \includegraphics[width=0.48\linewidth]{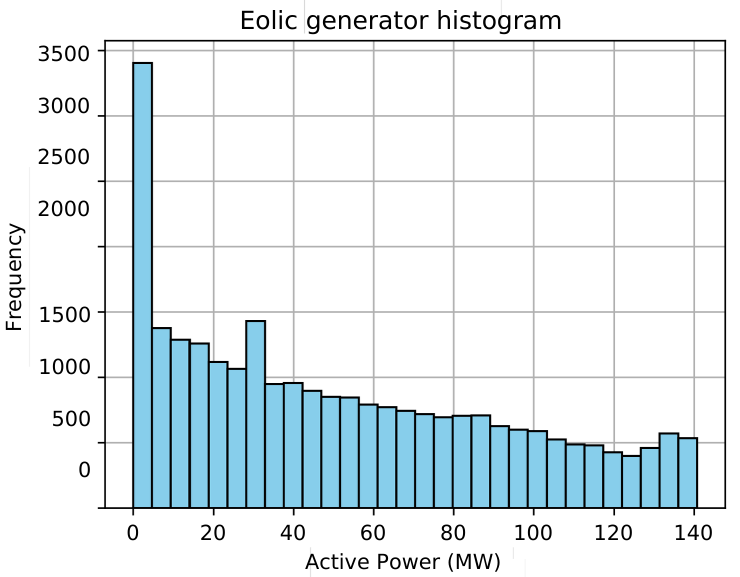}
        % \label{fig:wind_gen}
    % \end{subfigure}
    % \hfill
    % \begin{subfigure}{0.24\textwidth}
        \includegraphics[width=0.48\linewidth]{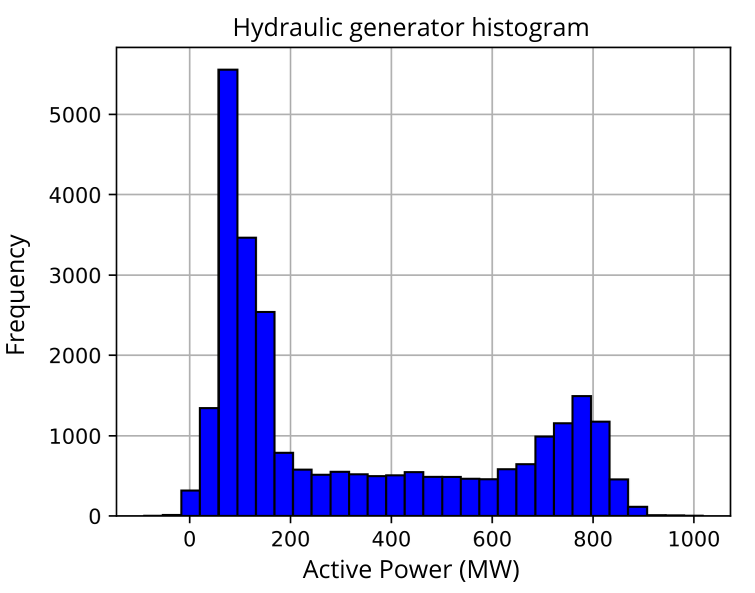}
        % \label{fig:hydro_gen}
    % \end{subfigure}
    % \begin{subfigure}{0.23\textwidth}
    %     \includegraphics[width=\linewidth]{images/solar.png}
    %     \caption{Solar generation.}
    %     \label{fig:solar_gen}
    % \end{subfigure}
    % \hfill
    % \begin{subfigure}{0.23\textwidth}
    %     \includegraphics[width=\linewidth]{images/biomasa.png}
    %     \caption{Biomass generation.}
    %     \label{fig:biomass_gen}
    % \end{subfigure}
    \caption{Histograms of power generation for different energy sources in the real dataset. They are far from uniform (as the synthetic case) and different between each other.}
    \label{fig:tipos_de_generacion}
\end{figure}

\noindent\textbf{Differences Between Real and Synthetic Datasets}.
Three key differences were identified between the real and synthetic datasets.
The first major difference lies in the complexity of the distributions across network components. In the synthetic dataset, generator outputs are sampled from a uniform distribution. In contrast, the real dataset exhibits significantly more intricate distributions that do not always conform to standard statistical models. \autoref{fig:tipos_de_generacion} illustrates this disparity by comparing the distribution of two randomly selected generators, one belonging to wind generation and the other to hydro.

% \begin{figure}
%     \centering
%     \begin{subfigure}{0.23\textwidth}
%         \includegraphics[width=\linewidth]{images/hist_gen_uru_sintetica_final.pdf}
%         \caption{Distribution of a synthetic generator.}
%         \label{fig:gen_sintetica}
%     \end{subfigure}
%     % \hfill
%     \begin{subfigure}{0.23\textwidth}
%         \includegraphics[width=\linewidth]{images/hist_real_provisorio.png}
%         \caption{Distribution of a real generator.}
%         \label{fig:gen_real}
%     \end{subfigure}
%     \caption{Comparison of generator output distributions in the synthetic and real datasets.}
%     \label{fig:primera_dif}
% \end{figure}

% \begin{figure}
%     \centering
%     \includegraphics[width=.3\textwidth]{images/top_loads.pdf}
%     \caption{Active power distribution of a random selected load in the real dataset.}
%     \label{fig:primera_dif}
% \end{figure}

The second difference is the variability in distribution types across different network elements in the real dataset. Unlike the synthetic dataset, where all generators follow a similar statistical pattern, real-world power grids consist of multiple generation types, such as wind, hydro, or solar, each with distinct statistical behaviors as also illustrated in \autoref{fig:tipos_de_generacion}.

\begin{figure}
    \centering
    % \begin{subfigure}{0.23\textwidth}
        \includegraphics[width=0.48\linewidth]{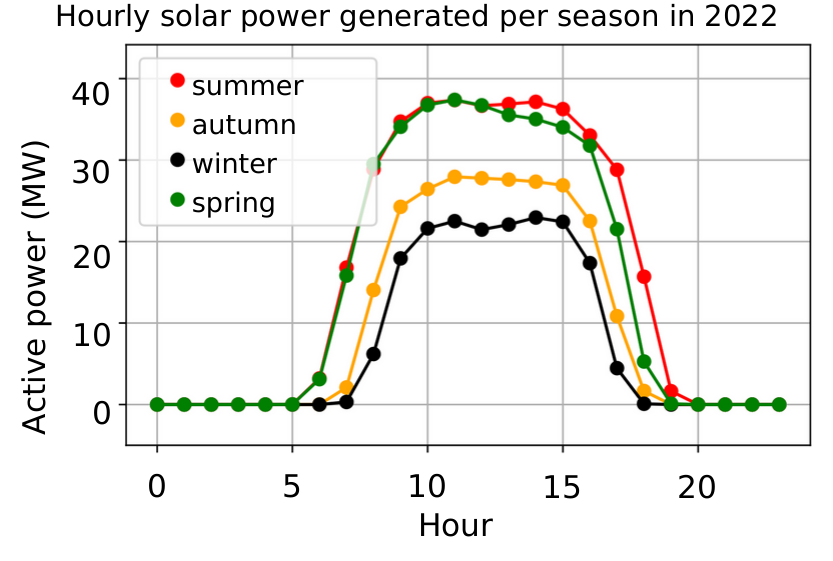}
        % \caption{Solar generation.}
        % \label{fig:wind_gen}
    % \end{subfigure}
    % \hfill
    % \begin{subfigure}{0.23\textwidth}
        \includegraphics[width=0.48\linewidth]{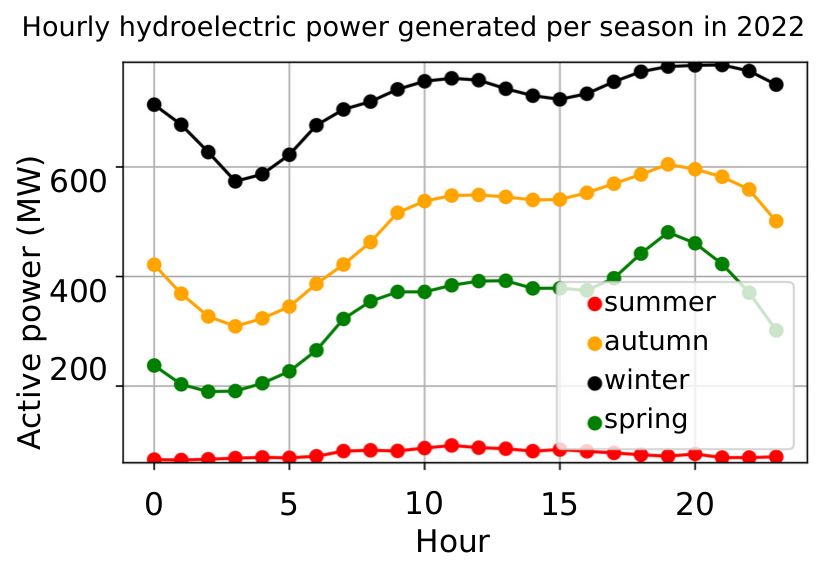}
        % \caption{Hydro generation.}
        % \label{fig:hydro_gen}
    % \end{subfigure}
    \caption{Average hourly renewable energy generation for different seasons in an year. Differently to the synthetic case, they exhibit seasonal patterns.}
    \label{fig:cambio_temporal}
\end{figure}

Finally, the real dataset exhibits distinct temporal and seasonal patterns that are absent in the synthetic dataset. For example, solar generators operate exclusively during daylight hours and remain inactive at night, unlike other generation types. Additionally, seasonal variations are evident: solar generation peaks in summer, while hydro generation tends to be more dominant in winter. \autoref{fig:cambio_temporal} illustrates these trends by depicting the average hourly generation across different seasons for solar and hydro power separately.

% \begin{figure}
%     \centering
%     \includegraphics[width=.5\textwidth]{images/gen_por_año.pdf}
%     \caption{Average hourly renewable energy generation for different seasons in an year. Differently to the synthetic case, they exhibit seasonal patterns. }

%     \label{fig:cambio_temporal}
% \end{figure}

% These differences underscore the challenges of using synthetic datasets to replicate real-world scenarios accurately. While synthetic data can be useful for benchmarking and controlled experiments, the absence of realistic distribution complexities, heterogeneity across components, and temporal dependencies may lead to models that perform well on synthetic data but fail to generalize effectively to real power grid operations.

\section{Experiments and Results}\label{sec:experiments}

% The implementation and evaluation of the proposed models are conducted using a publicly available repository on GitHub\footnote{https://github.com/tomyvazquez/DORAA-UY}, which contains the complete source code and configuration details. The datasets used for training and evaluation are stored in Google Drive \cite{drive_dataset} {\color{red} Cambiar al de la fing}, and all experiments are executed on an NVIDIA RTX 3060 GPU with 12GB of memory.

% To ensure a rigorous evaluation, the datasets are partitioned into training, validation, and test sets. Synthetic data is split randomly, while historical data is partitioned chronologically. Specifically, the training set includes data from January 2021 to April 2023 (78\%), the validation set covers April to September 2023 (14\%), and the test set spans September to December 2023 (8\%). Normalization is applied using statistics computed exclusively from the training set, ensuring consistent preprocessing and mitigating potential biases. 
Datasets were split into training, validation, and test sets. Synthetic data is partitioned randomly, while historical data follows a chronological split: January 2021–April 2023 (77\% training), April–September 2023 (18\% validation), and September–December 2023 (5\% test). %Normalization is applied using statistics from the training set to ensure consistency and prevent bias.
%
% Additionally, the data is organized into batches to optimize memory utilization, facilitate parallel processing, and improve model generalization. 
%
% The Mean Absolute Error (MAE) is adopted as an evaluation metric. Specifically, two distinct MAE formulations are considered: MAE$_v$ for voltage predictions at generator buses and MAE$_q$ for reactive power compensators. These metrics are computed as follows:
%
% \[
% \text{MAE}_\text{v} = \frac{1}{T}\sum_{t=1}^{T}\frac{||\mathbf{y}_t^{v}  - \Phi^{v}(\mathbf{x_t;\theta})||}{G}
% \]
%
% \[
% \text{MAE}_\text{q} = \frac{1}{T}\sum_{t=1}^{T}\frac{||\mathbf{y}_t^{q}  - \Phi^{q}(\mathbf{x_t;\theta})||}{R}
% \]
%
% where $\mathbf{y}_t^{v}$ and $\Phi^{v}(\mathbf{x_t;\theta})$ correspond to the voltage predictions at generator buses, while $\mathbf{y}_t^{q}$ and $\Phi^{q}(\mathbf{x_t;\theta})$ represent the reactive power control outputs. The terms $G$ and $R$ denote the number of voltage-controlling generators and reactive power compensators, respectively. By normalizing the absolute error over the number of controlled elements, these metrics provide a more interpretable assessment of model performance in predicting both voltage magnitudes and reactive power dispatch.
%
We retained only instances where pandapower's IPOPT converged (approximately 75\% of the total), which was not necessary in the synthetic case. This highlights the increased difficulty of optimizing real-world data, even for traditional methods.

The models incorporate regularization techniques such as weight decay, dropout, and early stopping to improve generalization and prevent overfitting. Additionally, hyperparameter optimization is performed using Optuna \cite{akiba2019optuna}. %, which applies Bayesian optimization to efficiently explore the search space and improve performance.  
Models are trained to simultaneously predict the voltage magnitudes at generator buses and the reactive power at compensators. Since the two variables have different scales, it is essential to normalize the outputs to prevent errors in reactive power predictions from dominating the loss function.
%, compared to errors in voltage magnitude predictions. 
Normalization is applied using statistics from the training set to ensure consistency and prevent bias.

The overall performance of each model is evaluated using the $\text{MAE}_v$ and $\text{MAE}_q$; i.e.\ the mean of the first and second column of $|\mathbf{Y}_t^*-\boldsymbol{\Phi}_{\boldsymbol{\theta}}(\mathbf{X}_t)|$, where again we only take into account the relevant entries at each node. Furthermore, 
for each input $\mathbf{X}_t$ and control variable $\mathbf{Y}_t$ (produced by either IPOPT or the learning algorithm) a power flow analysis is performed to compute the implicit variables and assess the total power loss $P_{\text{loss}}$ and constraint violations. For each instance of the problem $t$ we compute the resulting $P_{\text{loss}}$ of the learning methods and the relative difference with the value obtained by IPOPT (we report the mean $\pm$ the standard deviation in the test set), and if all constraints are enforced (we report the percentage of instances where all constraints are enforced). 
Results for both synthetic and historical data are presented in Table \ref{tab:comparison}, which we now discuss.

% \begin{table}
% \centering
% \scriptsize
% \setlength{\tabcolsep}{3pt} % Reduce column separation
% \renewcommand{\arraystretch}{1.1} % Slightly compress row height
% \caption{Comparison of network losses, feasibility, and prediction errors for models using synthetic and historical data. Feas.* shows feasible solutions after a 1.8\% constraint relaxation.}
% \begin{tabular}{lcccccc}
% \toprule
%  & Model & Losses & Feas. & Feas.* & $\text{MAE}_v$ & $\text{MAE}_q$ \\
%  &       & (MW)  & (\%)  & ($1.8\%$) & & \\ \midrule
% \multicolumn{7}{l}{\textbf{Synthetic Data}} \\ \midrule
%  & Optimal & 17.53 & 100 & -- & -- & -- \\
%  & FCNN    & 17.53 & 98.1 & -- & $4.6 \times 10^{-4}$ & $2.8 \times 10^{-1}$ \\
%  & GNN     & 17.53 & 98.6 & -- & $5.5 \times 10^{-4}$ & $2.9 \times 10^{-1}$ \\ \midrule
% \multicolumn{7}{l}{\textbf{Historical Data}} \\ \midrule
%  & Optimal & 28.3 & 79.8 & 100 & -- & -- \\
%  & FCNN    & 28.3 & 64.8 & 92.6 & $2.6 \times 10^{-3}$ & 2.75 \\
%  & GNN     & 28.3 & 61.4 & 74.6 & $3.7 \times 10^{-3}$ & 3.76 \\ \bottomrule
% \end{tabular}

% \label{tab:comparison}
% \end{table}

\begin{table}
\centering
\scriptsize
\setlength{\tabcolsep}{3pt} % Reduce column separation
\renewcommand{\arraystretch}{1.1} % Slightly compress row height
\caption{Comparison of network losses, feasibility, and prediction errors for models using synthetic and historical data. Feas.* shows feasible solutions after a 1.8\% constraint relaxation.}
\begin{tabular}{lcccccc}
\toprule
 & Model & Losses & Feas. & Feas.* & $\text{MAE}_v$ & $\text{MAE}_q$ \\
 &       & [\% relative to IPOPT]  & \multicolumn{2}{c}{[\% of instances]} & [p.u.] & [MVar] \\ \midrule
\multicolumn{7}{l}{\textbf{Synthetic Data}} \\ \midrule
 & Optimal & $0\pm 0$ & 100 & -- & -- & -- \\
 & FCNN    & $0.23\pm 0.63$ & 98.1 & -- & $4.6 \times 10^{-4}$ & $2.8 \times 10^{-1}$ \\
 & GNN     & $0.12\pm 0.66$ & 98.6 & -- & $5.5 \times 10^{-4}$ & $2.9 \times 10^{-1}$ \\ \midrule
\multicolumn{7}{l}{\textbf{Historical Data}} \\ \midrule
 & Optimal & $0\pm 0$ & 79.8 & 100 & -- & -- \\
 & FCNN    & $0.00\pm 0.69$ & 64.8 & 92.6 & $2.6 \times 10^{-3}$ & 2.75 \\
 & GNN     & $-0.25\pm 0.70$ & 61.4 & 74.6 & $3.7 \times 10^{-3}$ & 3.76 \\ \bottomrule
\end{tabular}

\label{tab:comparison}
\end{table}

% fcnn - sintéticos
% mean: 2.27E-03; std: 6.27E-03
% gnn - sintéticos
% mean: 1.20E-03; std: 6.56E-03
% fcnn - históricos
% mean: 3.33E-16; std: 6.93E-03
% gnn - históricos
% mean: -2.48E-03; std: 6.98E-03

%%% en porcentaje (o sea *100)
% fcnn - sintéticos
% mean: 0.23; std: 0.63
% gnn - sintéticos
% mean: 0.12; std: 0.66
% fcnn - históricos
% mean: 0.00; std: 0.69
% gnn - históricos
% mean: -0.25; std: 0.70

% \subsection{Synthetic Data}
% The best-performing architectures achieve voltage prediction errors of $\text{MAE}_v=4.6 \times 10^{-4}$ for FCNN and $\text{MAE}_v=5.5 \times 10^{-4}$ for GNN, while the reactive power prediction errors are $\text{MAE}_q=2.8 \times 10^{-1}$ and $\text{MAE}_q=2.9 \times 10^{-1}$, respectively.

\begin{figure}
    \centering
    \includegraphics[width=.45\textwidth]{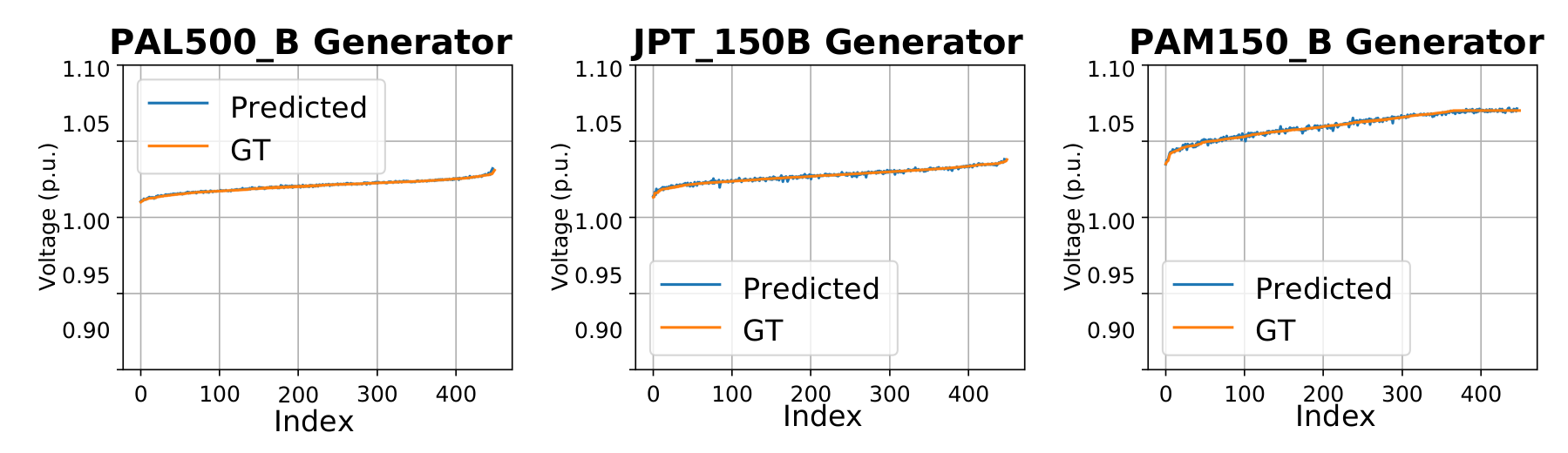}
    \includegraphics[width=.45\textwidth]{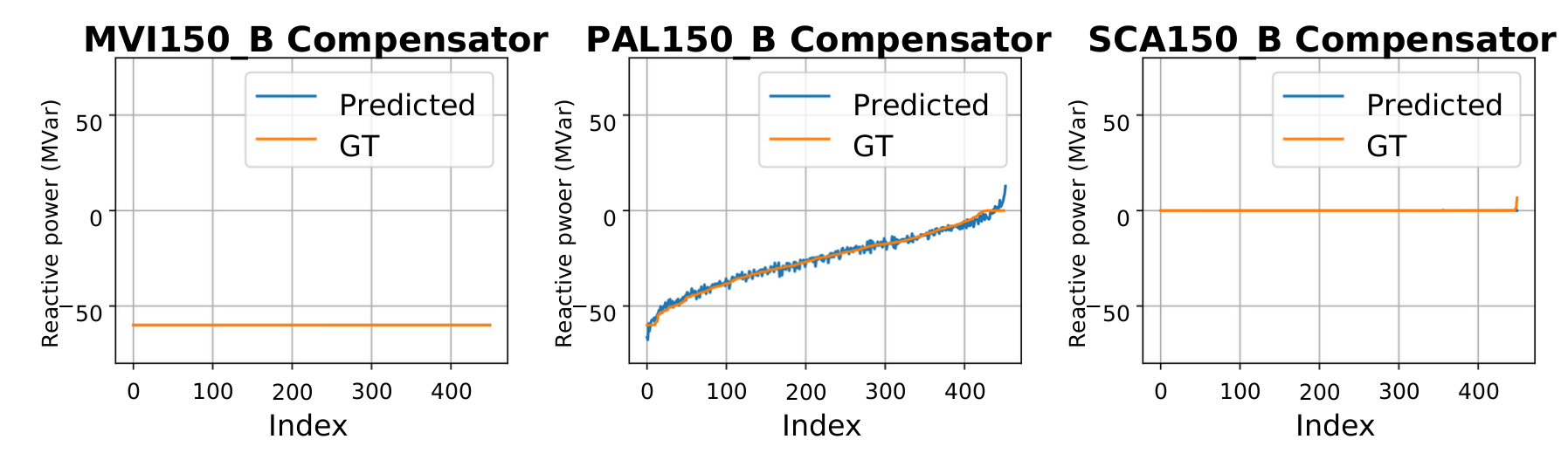}
    \caption{Comparison between optimal and predicted voltages (above) and reactive power (below) for the GNN in the Uruguayan network with synthetic data. Variations are not significant and the learning method obtains great performance.}
    \label{fig:qshunts_GNN}
\end{figure}

\noindent\textbf{Synthetic Data.}
% Figure \ref{fig:qshunts_GNN} compares the optimal and predicted values for voltage and reactive power using the GNN model. This example considers three randomly chosen generators and compensators in the electrical network to illustrate the model's performance. The results for the FCNN model are practically identical and both models effectively approximate the optimizer's solutions with high accuracy, as shown in Table \ref{tab:comparison}. The obtained voltages and reactive powers are very similar to the optimal, resulting in a total loss that is always within 2\% the one obtained by IPOPT. Furthermore, in the vast majority of the validation set instances, all constraints were enforced. A more fine analysis reveals that in these constraint violations the infringing variables were always within 5\% of the corresponding interval's length.
%
Figure \ref{fig:qshunts_GNN} compares the optimal and predicted voltage and reactive power values obtained using the GNN model. To illustrate the model's performance, we consider three randomly selected generators and compensators within the network, and show the ground-truth and predicted control variables over the test set.
For better visualization, we reorder $t$ so that the ground-truth values increase monotonically, denoted as ``Index'' on the x-axis. 
The results for the FCNN model are nearly identical, with both models closely approximating the optimizer’s solutions, as quantified in Table \ref{tab:comparison}. The predicted voltages and reactive power values closely match the optimal ones, yielding total losses that remain within 2\% of those obtained by IPOPT. Additionally, constraints were satisfied in the vast majority of validation instances. A more detailed analysis shows that when violations occurred, the affected variables deviated by at most 5\% of their respective constraint bounds, indicating minimal constraint breaches.

% \begin{figure}
%     \centering
%     \includegraphics[width=.45\textwidth]{images/valores_reales_predecidos_FCNN_global_vmpu_sintetica_en-cropped.pdf}
%     \caption{Comparison between optimal and predicted voltages for GNN in the Uruguayan network with synthetic data.}
%     \label{fig:volts_GNN}
% \end{figure}

% \begin{figure}
%     \centering
%     \includegraphics[width=.45\textwidth]{images/valores_reales_predecidos_FCNN_global_qshunts_sintetica_en-cropped.pdf}
%     \caption{Comparison between optimal and predicted reactive power for GNN in the Uruguayan network with synthetic data.}
%     \label{fig:qshunts_GNN}
% \end{figure}

% Although the models closely replicate the optimal solutions, the primary objective is to solve the Optimal Reactive Power Dispatch (ORPD) problem. For each test sample $\mathbf{x}^t$, the predicted control variables $\Phi(\mathbf{x}^t, \theta)$ are applied to the network, and a power flow analysis is performed to assess system losses and constraint violations. These results are compared with the optimal solution.

% Table \ref{tab:comparison} summarizes the comparison of different strategies for solving the ORPD, presenting the average network losses and the percentage of feasible test cases (i.e., cases without constraint violations). Both learning-based models achieve feasibility rates above $98\%$, demonstrating their effectiveness in generating solutions that closely match the optimal ones.

% \begin{table}[h]
%     \centering
%     \caption{Comparison of network losses and feasibility for best-performing models with synthetic data.}
%     \begin{tabular}{ccc}
%         \hline
%         & Losses (MW) & Feasibility \\ \hline
%         Optimal & $17.53$ & $100\%$ \\
%         FCNN & $17.53$ & $98.1\%$ \\
%         GNN & $17.53$ & $98.6\%$ \\
%     \end{tabular}
    
%     \label{tab:comparison}
% \end{table}

\begin{figure}
    \centering
    \includegraphics[width=.48\textwidth]{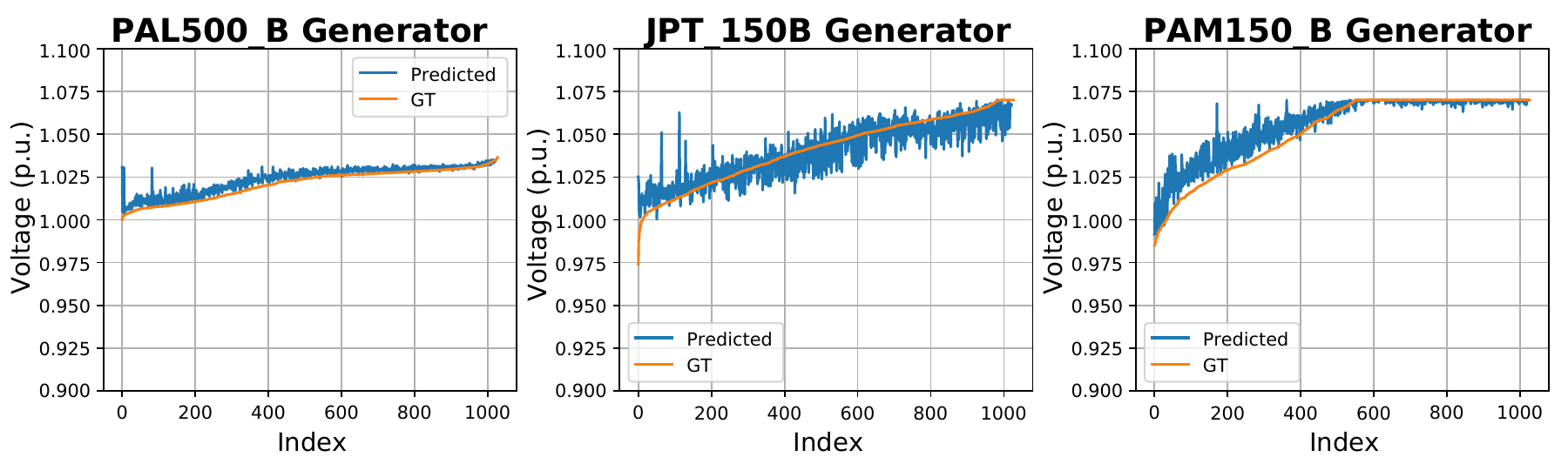}
    \includegraphics[width=.48\textwidth]{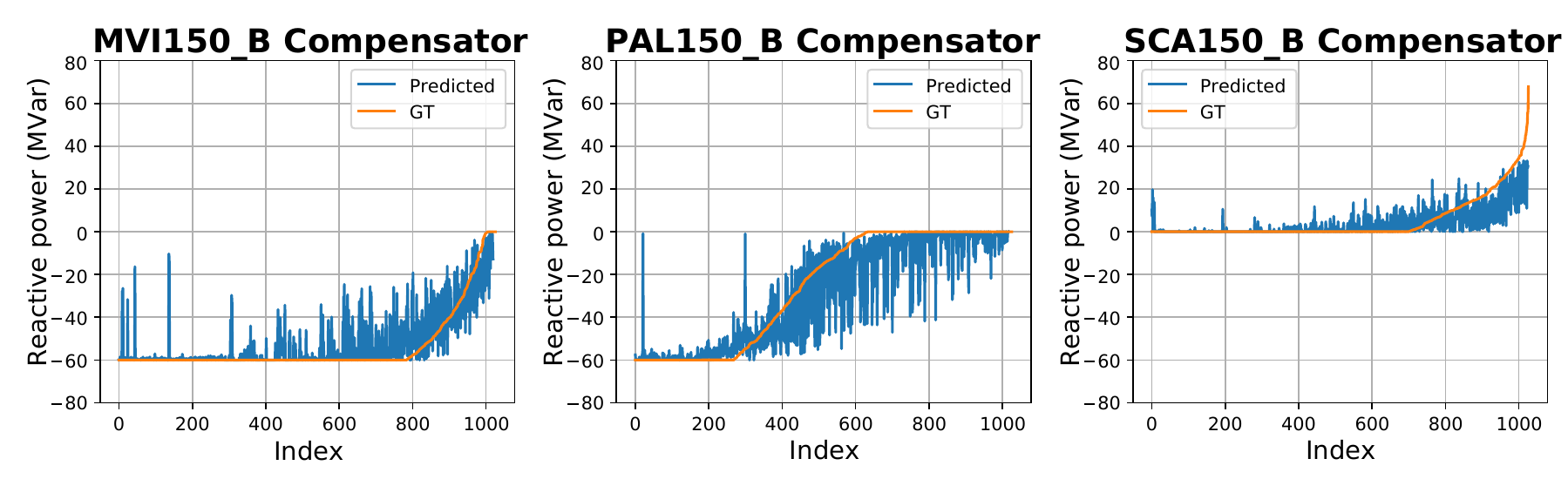}
    \caption{Comparison between optimal and predicted voltages (above) and reactive power in compensators (below) for the GNN with historical data. Variations are significant and the learning method struggles to follow the optimal values.}
    \label{qshunts_supuru_GNN}
\end{figure}

\noindent\textbf{Historical Data.}
We now discuss the performance of the learning strategies when trained and evaluated using the actual historical data (see the lower part of Table \ref{tab:comparison}). 
% The best-performing architectures achieve MAE for voltage prediction $\text{MAE}_v=2.6\times10^{-3}$ for FCNN and $\text{MAE}_v=3.7\times10^{-3}$ for GNN, while the MAE for reactive power prediction is $\text{MAE}_q=2.75$ and $\text{MAE}_q=3.76$, respectively. 
Note that although the resulting total loss is still very close to the value obtained by IPOPT, both $\text{MAE}_v$ and $\text{MAE}_q$ are approximately 10 times higher than those obtained in the synthetic case. A comparison between the predicted and optimal values for some randomly chosen nodes is presented in \autoref{qshunts_supuru_GNN}. Not only are the predictions significantly noisier and further from the optimal values, but the former exhibits a broader range of possible values. This increased variability emphasizes the additional challenges introduced by real-world distributions for predictive models.

% \begin{figure}[h]
%     \centering
%     \includegraphics[width=.45\textwidth]{images/valores_reales_predecidos_GNN_local_vm_pu_opt.pdf-cropped.pdf}
%     \caption{Comparison between optimal and predicted voltages using GNN for historical data.}
%     \label{volts_supuru_GNN}
% \end{figure}

% \begin{figure}[h]
%     \centering
%     \includegraphics[width=.45\textwidth]{images/valores_reales_predecidos_GNN_local_qshunts.pdf-cropped.pdf}
%     \caption{Comparison between optimal and predicted reactive power in compensators using GNN for historical data.}
%     \label{qshunts_supuru_GNN}
% \end{figure}

Regarding feasibility, \autoref{tab:comparison} shows that even the IPOPT of PandaPower does not enforce all constraints, and some very slightly exceed the prescribed limits. This finding further demonstrates that the problem is inherently challenging, not only for predictive models but also for traditional optimization techniques. Naturally, both learning methods also fail to enforce all constraints in a significant portion of the test cases. 
% presents the feasibility and loss comparisons among different models. Notably, the optimal model does not reach 100\% feasibility because some constraints slightly exceed the prescribed limits. This finding demonstrates that the problem is inherently challenging, not only for predictive models but also for traditional optimization techniques.
%
In particular, by increasing all constraints by just 1.8\%, solutions obtained by PandaPower enforce all constraints. In this case, the FCNN still fails to enforce all constraints in about 10\% of the instances, whereas the GNN in 25\%, a very poor performance. % When adjusting the constraints to ensure that the optimizer achieves 100\% feasibility, the learnable models exhibit improved results. Specifically, FCNN reaches a feasibility of 92.6\%, whereas GNN achieves 74.6\%. However, these values are still lower than those observed in the synthetic case, where feasibility metrics were considerably higher.

% \begin{table}[h]
% \centering
% \caption{Comparison of network losses and feasibility for best-performing models using historical data. The Feasibility* column represents the percentage of feasible solutions after relaxing constraints just enough to ensure 100\% feasibility for the optimizer, which in this case corresponds to a $1.8\%$ relaxation.}
% \begin{tabular}{cccc}
% \hline
%             & Losses (MW)     & Feasibility   & Feasibility* ($1.8\%$)   \\ \hline
% Optimal      & $28.3$            &  $79.8\%$  &  $100\%$   \\
% FCNN        & $28.3$            & $64.8\%$    & $92.6\%$         \\
% GNN         & $28.3$            & $61.4\%$    & $74.6\%$     
% \end{tabular}
% \label{supuru_comparacion}
% \end{table}

\section{Conclusions}

We have presented a new publicly available dataset that includes a real-world electrical network along with almost two years' worth of demand and generation. Using this dataset, we evaluated two learning-based optimizers inspired by~\cite{owerko2020optimal}, which learn to solve the ORPD problem by imitating IPOPT solutions. While these models perform exceptionally well on synthetic data—where demands and generation are uniformly sampled—their accuracy deteriorates significantly when applied to real-world data. This degradation leads to constraint violations in several cases. Similar trends were observed with the unsupervised approach proposed in~\cite{owerko2024unsupervised}, though those results are omitted here due to space constraints.

These findings underscore the gap between synthetic and real-world scenarios. Factors such as seasonal variations and the complex, non-standard distributions of demand and generation introduce additional challenges that hinder learning and reduce prediction accuracy. As future work, it would be valuable to explore constraint-enforcing techniques, such as those in~\cite{pan2023deepopf}, extend them to the ORPD problem, and assess their effectiveness using our dataset.

% These results highlight the contrast between the synthetic and real-world cases. 
% % The use of real-world data significantly impacts the predictor performance due to its increased complexity. 
% % Additionally, discrepancies between training, validation, and test distributions, caused by temporal dependencies, further challenge model generalization. 
% Seasonal variations, and different non-standard distributions of demands and generated power are factors that contribute to
% % significant fluctuations in power demand, 
% complicating the learning process and reducing prediction accuracy. 

\bibliographystyle{IEEEtran}
\bibliography{referencias}

\end{document}